\documentclass[conference]{IEEEtran}
\IEEEoverridecommandlockouts
\usepackage{cite}
\usepackage{amsmath,amssymb,amsfonts}
\usepackage{graphicx}
\usepackage{textcomp}
\usepackage{xcolor}
\usepackage{enumitem}

\usepackage[noend]{algpseudocode}

\usepackage{amsthm}
\usepackage{subcaption}
\newcommand{\ve}[1]{{\mbox{\boldmath${#1}$}}}
\theoremstyle{definition}

\usepackage{balance}
\newcommand{\norm}[1]{\left\lVert#1\right\rVert}
\newcommand{\floor}[1]{\lfloor #1 \rfloor}

\def\BibTeX{{\rm B\kern-.05em{\sc i\kern-.025em b}\kern-.08em
    T\kern-.1667em\lower.7ex\hbox{E}\kern-.125emX}}
\begin{document}

\title{Spatio-Temporal Functional Neural Networks}

\author{
    \IEEEauthorblockN{Aniruddha Rajendra Rao\IEEEauthorrefmark{2}$^{1}$\thanks{$^{1}$With equal contributions. This work has been conducted during Aniruddha Rajendra Rao’s internship at Hitachi America, Ltd.} , Qiyao Wang\IEEEauthorrefmark{1}$^{1}$, Haiyan Wang\IEEEauthorrefmark{1}, Hamed Khorasgani\IEEEauthorrefmark{1}, Chetan Gupta\IEEEauthorrefmark{1}}
   \IEEEauthorblockA{\IEEEauthorrefmark{2}Department of Statistics, Penn State University, University Park, PA
   \\\{arr30\}@psu.edu}
   \IEEEauthorblockA{\IEEEauthorrefmark{1}Industrial AI Lab, Hitachi America, Ltd. R$\&$D, Santa Clara, CA
    \\\{Qiyao.Wang, Haiyan.Wang, Hamed.Khorasgani, Chetan.Gupta\}@hal.hitachi.com}
}

\maketitle
\footnotetext[2]{\copyright 2020 IEEE. Personal use of this material is permitted.  Permission from IEEE must be obtained for all other uses, in any current or future media, including reprinting/republishing this material for advertising or promotional purposes, creating new collective works, for resale or redistribution to servers or lists, or reuse of any copyrighted component of this work in other works.}

\begin{abstract}
Explosive growth in spatio-temporal data and its wide range of applications have attracted increasing interests of researchers in the statistical and machine learning fields. The spatio-temporal regression problem is of paramount importance from both the methodology development and real-world application perspectives. Given the observed spatially encoded time series covariates and real-valued response data samples, the goal of spatio-temporal regression is to leverage the temporal and spatial dependencies to build a mapping from covariates to response with minimized prediction error. Prior arts, including the convolutional Long Short-Term Memory (CovLSTM) and variations of the functional linear models, cannot learn the spatio-temporal information in a simple and efficient format for proper model building. In this work, we propose two novel extensions of the Functional Neural Network (FNN), a temporal regression model whose effectiveness and superior performance over alternative sequential models have been proven by many researchers. The effectiveness of the proposed spatio-temporal FNNs in handling varying spatial correlations is demonstrated in comprehensive simulation studies. The proposed models are then deployed to solve a practical and challenging precipitation prediction problem in the meteorology field. 
\end{abstract}


\section{Introduction}\label{sec1}
Nowadays, spatially correlated data has grown at an explosive rate in domains such as meteorology, epidemiology, transportation, agriculture, industrial networks (e.g., power grids and water supply networks), and social science \cite{yu2017spatio, rossi2018relational, steiger2015advanced, tastu2011spatio, oliveira2019biased}. Also, most spatial data naturally evolves dynamically over time. For instance, weather stations scattered over a geographic area continuously monitor the location-specific weather conditions over time. This type of data that encodes both spatial and temporal dependencies is called the spatio-temporal data \cite{cressie2015statistics, bailey1995interactive}. Conventional statistical and machine learning models that assume independently and identically distributed (i.i.d) data, time series analysis that ignores the spatial correlations, and spatial data analysis that ignores the temporal dependencies are often not suitable for analyzing spatio-temporal data. Comprehensive reviews for research and technology developed specially for this emerging type of data can be found in \cite{shekhar2015spatiotemporal, roddick1999bibliography, roddick2000updated}.

\begin{figure}
\centering
\includegraphics[width=90mm]{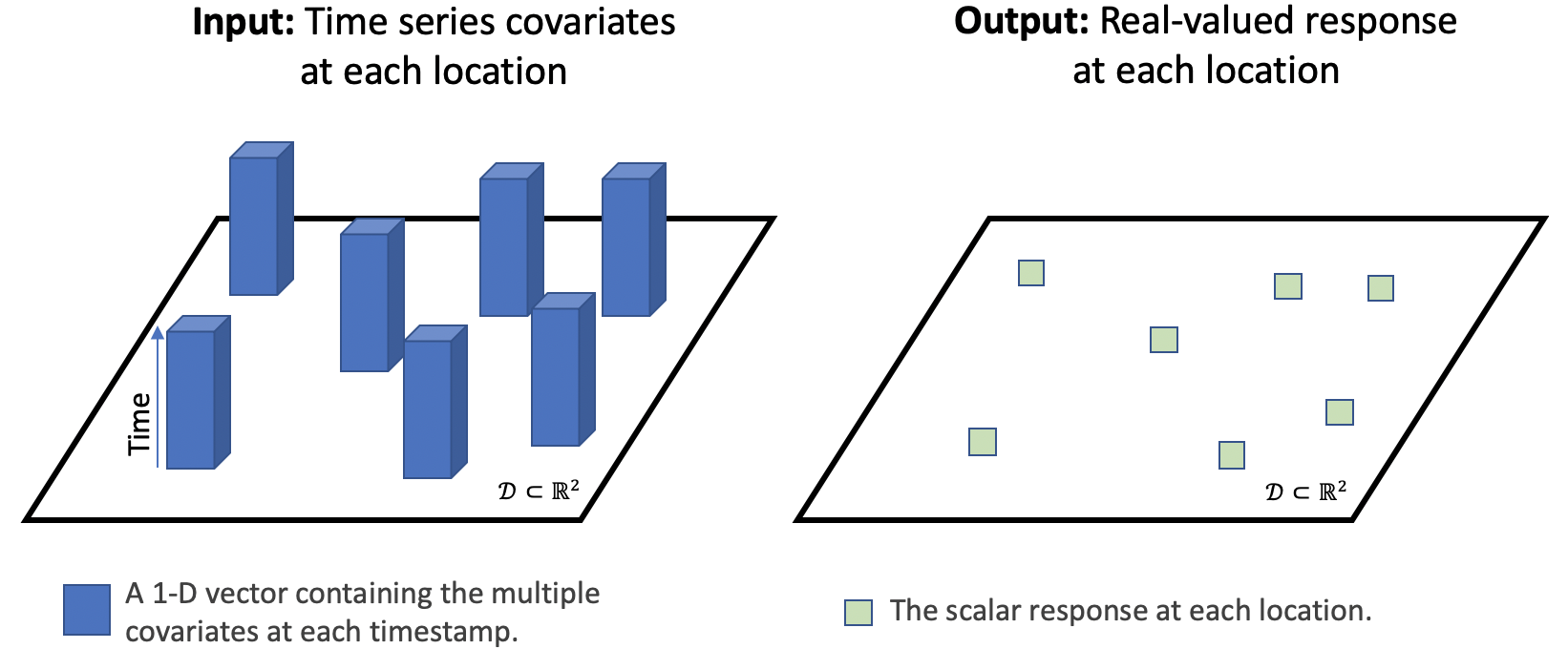}
  \caption{Illustration of input and output data in spatio-temporal regression.}
  \label{data}
\end{figure}

This article specifically focus on the \textit{spatio-temporal regression} problem, one of the most fundamental problems in the diverse spatio-temporal research area \cite{shekhar2015spatiotemporal}. In particular, given a set of spatially correlated time series, the task is to estimate the real-valued response variable for each time series input. An example of spatio-temporal input and output data is illustrated in Fig \ref{data}. Solutions with high prediction accuracies have profound impacts on a wide range of applications. For instance, for an industrial network consists of multiple interacting pieces of equipment that are usually geographically dispersed such as power grids and water supply networks, valid spatio-temporal regression models significantly improve maintenance efficiency by precisely predicting the remaining time to the next failure for each equipment based on equipment condition monitoring data up to the present time. The considered regression problem is non-trivial. Compared with the temporal regression that assumes samples are independent, there are at least two additional dimensions of complexity in the underlying relationship: 1) \textit{Heterogeneity:} The relationship may be location-specific \cite{yamanishi20038}; 2) \textit{Dependency:} The relationship at one place may be affected by other locations \cite{huang2018spatial}.

One promising and straightforward way of incorporating the spatial aspect is to treat the spatial information as an additional scalar covariate for the temporal regression models, e.g., the Recurrent Neural Network (RNN) \cite{jordan1990attractor}, the Long Short-Term Memory networks (LSTM) \cite{hochreiter1997long}, and the Functional Neural Network (FNN) \cite{rossi2002functional, wang2019multilayer}. However, this strategy usually discards the rich spatial information contained in the data and is inadequate to capture the spatial effect on the underlying relationship. 

More sophisticated spatio-temporal regression models have been developed in the literature. In the deep learning community, efforts have been taken to combine the Convolutional Neural Network (CNN) specialized in handling image data and the baseline deep learning models for temporal data, such as the RNN and LSTM, to build powerful predictive models with spatio-temporal data \cite{xingjian2015convolutional, baccouche2011sequential, jiang2017supercharging}. For instance, a convolutional LSTM (ConvLSTM) \cite{xingjian2015convolutional} has been proposed to use the previously recorded radar echo sequence to forecast the future radar maps in the considered region. However, these models essentially learn a mathematical mapping from the 3-D tensor input (indexed by time and 2-D space) to a 2-D response (indexed by space) and therefore require a large number of the input/output data pairs shown in Fig \ref{data} to fit the unknown parameters. For instance, in one of the experiments in \cite{xingjian2015convolutional}, the ConvLSTM model was learned with 10000 input/output training pairs and 2000 validation data pairs. As only one input/output instance needed in such models is accessible, the deep learning models that encode spatial information through convolutional layers are \textit{not applicable to} the considered spatio-temporal prediction problem illustrated in Fig \ref{data}. 

In the statistical field, Functional Linear Regression (FLM) is a standard predictive model with i.i.d time series input and real-valued scalar response \cite{cardot1999functional, yao2005functional}. Two extensions of FLM have been proposed to account for the spatial heterogeneity and dependency, with the help of the geographically weighted regression and the spatial autoregression techniques respectively \cite{yamanishi20038, huang2018spatial}. The former statistical trick enables the building of a location-specific temporal functional regression model. The latter one explicitly incorporates the impact from nearby locations into the i.i.d FLM. Unlike the convolutional layer-based models, the generalized functional regression models treat the time series covariates from different locations as individual random samples, which makes them suitable solutions to learn the relationship between the spatially correlated temporal inputs and scalar targets in Fig \ref{data}. However, these models are linear and therefore suffer from underfitting when the underlying mapping is complex.

To overcome the limitations of the linear models, leveraging the geographically weighted regression and the spatial autoregression techniques, we propose two novel models to generalize the non-linear model Functional Neural Networks (FNN) for the spatial-temporal regression problem. It is noteworthy that the effectiveness of FNN and its superior performance over the state-of-art i.i.d temporal regression models (e.g., MLP, RNN, LSTM, CNN) for some real-life problems have been demonstrated by previous research work \cite{rossi2002functional, wang2019remaining, wang2019multilayer}. The contributions of this paper are summarized as follows:
\begin{enumerate}[leftmargin=*]
    \item We propose two novel models (Geographically Weighted FNN in Section \ref{sec3} and Spatial Autoregressive FNN in Section \ref{sec4}) to respectively handle the spatial heterogeneity and dependency in spatio-temporal regression problems.  
    \item For each model, we propose an effective way to predict the response at new spatial locations (see later discussions).
    \item We conduct simulation studies to justify the effectiveness of our proposed models in varying situations. We apply the models to solve a real-world challenge, i.e., the critical precipitation prediction problem in meteorology.  
\end{enumerate}

\section{Preliminaries}\label{sec2}

\subsection{Formulation of Spatio-temporal Prediction}\label{sec2.1}
The goal of spatio-temporal regression is to build a mapping from time series covariates to a real-valued scalar response, leveraging both the spatial and temporal dependencies among data samples to minimize the prediction error. 

Suppose that we observe $N$ samples scattered over a $d$-dimensional spatial region $\mathcal{D} \subset \mathbb{R}^d$,  $d\in \mathbb{Z}^+$. For each subject $i\in \{1,2,...,N\}$, $R$ covariates are continuously recorded within a compact time interval $\mathcal{T}\subseteq \mathbb{R}$. Note that the measuring timestamps can vary across different covariates and different data samples. Hence, the subject and feature indexes need to be reflected in the following notations. In particular, the measuring times of the $r$-th feature of subject $i$ are denoted as $\mathbf{T^{(i,r)}}=[T^{(i,r)}_{1},...,T^{(i,r)}_{j},...,T^{(i,r)}_{M^{(i,r)}}]^T$, with $M^{(i,r)}$ being the number of observations and the measuring timestamps $T^{(i,r)}_{j}\in \mathcal{T}$ for $i=1,...,N; r=1,...,R; j=1,...,M^{(i,r)}$. The corresponding covariate data are represented by $\mathbf{X^{(i,r)}}=[X^{(i,r)}_{1},...,X^{(i,r)}_{j},...,X^{(i,r)}_{M^{(i,r)}}]^T$. The real-valued response variable for the $i$-th subject is $Y_i \in \mathbb{R}$ and the spatial information is presented by $\mathbf{S_i}\in\mathcal{D}$. In summary, the observed data is $\{\mathbf{X^{(i,1)}}, ..., \mathbf{X^{(i,R)}}, Y_i, \mathbf{S_i}\}_{i=1}^N$.

For the traditional temporal regression problem, the goal is to use i.i.d samples $\{\mathbf{X^{(i,1)}}, ..., \mathbf{X^{(i,R)}}, Y_i\}_{i=1}^N$ to learn a singular mapping 
\begin{equation} \label{e.iid}
Y_i = F(\mathbf{X^{(i,1)}}, ..., \mathbf{X^{(i,R)}}, \ve{\theta}),
\end{equation}
where $\ve{\theta}$ represents the unknown population parameters that apply uniformly to any instance $i$. When additional spatial information and dependencies are encoded into the input/output data, the regression problem is typically formulated as learning the location specific mapping in Eq (\ref{e1.22}) or the spatial autoregressive mapping in Eq (\ref{e1.33}) \cite{yamanishi20038, huang2018spatial}.

\begin{equation} \label{e1.22}
Y_i = F(\mathbf{X^{(i,1)}}, ..., \mathbf{X^{(i,R)}}, \ve{\theta}_{\mathbf{S_i}})
\end{equation}

\begin{equation} \label{e1.33}
Y_i = F(\mathbf{X^{(i,1)}}, ..., \mathbf{X^{(i,R)}}, \mathbf{Y}, \ve{\tilde{\theta}})
\end{equation}
In Eq (\ref{e1.22}), $\ve{\theta}_{\mathbf{S_i}}$ presents the unknown location-specific parameters. In Eq (\ref{e1.33}), the common parameters $\ve{\tilde{\theta}}$ contains both the temporal predictor's parameter $\ve{\theta}$ and the spatial autoregressive parameters that specify the relationship among response variables at different spatial locations $\mathbf{Y}=[Y_1,...,Y_N]^T$. The prior art \cite{yamanishi20038, huang2018spatial} respectively learns a linear formatted function $F(\cdot)$. Whereas, our proposed models in Section \ref{sec3} and \ref{sec4} can capture more complex relationships.

\subsection{Functional Data Analysis and Functional Neural Network}\label{sec2.2}
Functional data analysis (FDA) is a rapidly growing branch of statistics specialized in representing and modeling dynamically varying data over a continuum (e.g., time) \cite{ramsay2006functional,Reimherr2017FDA} . FDA models uniquely treat $\mathbf{X^{(i,r)}}$, the observed $r$-th feature of subject $i$ over time, as discretized observations from a continuous underlying curve $X^{(i,r)}(t), t \in  \mathcal{T}$. Under the regression setting, the conventional Functional Linear Models (FLM) \cite{cardot1999functional, yao2005functional} assume and learn the unknown real-valued parameters in the mapping 
\begin{equation} \label{flm}
    Y_i = b + \sum_{r=1}^{R} \int_{t\in \mathcal{T}}W_{r}(\ve{\beta}_{r},t)X^{(i,r)}(t)dt,
\end{equation}
where $b\in \mathbb{R}$ is the unknown intercept, and $\ve{\beta}_{r}$ is a finite-dimensional vector that quantifies the parameter function $W_{r}(\ve{\beta}_{r},t)$, for $r=1,...,R$. As FLMs directly handles the continuous underlying random processes $\{X^{(i,r)}(t)\}_{r=1}^R$ and the unknown weight functions $W_{r}(\ve{\beta}_{r},t)$ are also continuous functions over time, FLMs possess several advantages over the alternative temporal regression models such as RNN and LSTM: 1) FLMs are capable of handling versatile data formats. In particular, the time series covariates can be regular or irregular. Also, the number of observations as well as the measuring timestamps can be different across features and across subjects \cite{yao2005functional}. 2) FLMs effectively capture the timely varying correlation between the covariates and the response, while the sequential deep learning models typically use the same set of parameters across all timestamps \cite{ramsay2006functional,Reimherr2017FDA}.

Functional neural network (FNN) is introduced by \cite{rossi2002functional} and later investigated further by \cite{wang2019remaining, wang2019multilayer}, with the purpose of learning complex mappings between functional covariates $\{X^{(i,r)}(t)\}_{r=1}^R$ and scalar responses $Y_i$ from i.i.d data samples. FNNs have demonstrated improved performances over the alternative methods (e.g., RNN, LSTM) in temporal to scalar predictive modeling tasks, including prediction of patient's long-term survival using their serum bilirubin measurements within the first month of the treatment, and equipment's remaining useful life  estimations using historical sensor time series \cite{rossi2002functional, wang2019remaining, wang2019multilayer}. 

\begin{figure}
\centering
\includegraphics[width=85mm]{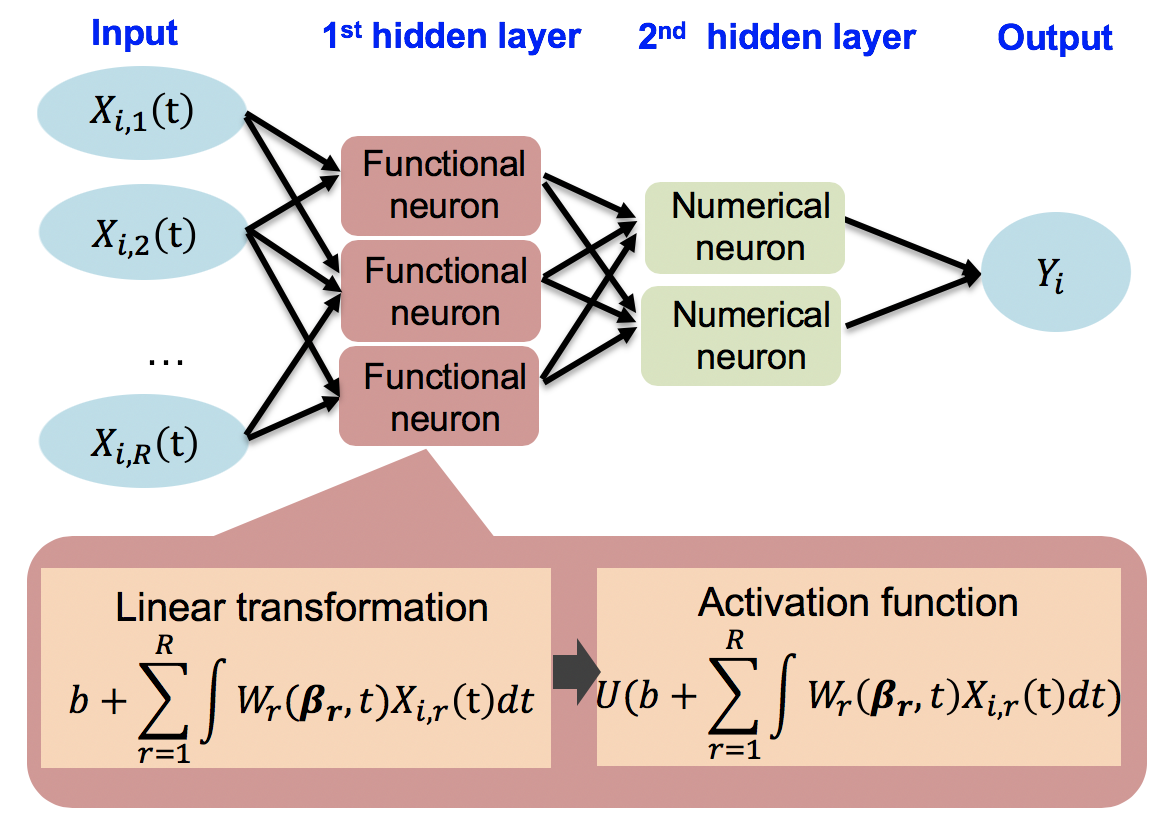}
  \caption{The architecture of a FNN with three functional neurons on the first layer and two numerical neurons on the second layer.}
  \label{FNN_exp}
\end{figure}

The fundamental idea of FNN is to embed the FLM in Eq (\ref{flm}) into the fully connected neural network structure in deep learning \cite{rossi2002functional}. In particular, the architecture of FNN is described as follows. The first layer of FNN consists of novel \textit{functional neurons}. The functional neurons take the functional covariates $\mathbf{X_i(t)}=[X^{(i,1)}(t),...,X^{(i,R)}(t)]^T$ as input and calculate
\begin{equation} \label{e2.2.1}
H(\mathbf{X_i(t)}, \ve{\beta})=U(b + \sum_{r=1}^{R} \int_{t\in \mathcal{T}}W_{r}(\ve{\beta}_{r},t)X^{(i,r)}(t)dt), 
\end{equation}
where $b$ and $W_{r}(\ve{\beta}_{r},t)$ are the same as those in Eq (\ref{flm}), $\ve{\beta}=[\ve{\beta}_1,...,\ve{\beta}_R]^T$, and $U(\cdot)$ is a nonlinear activation function from $\mathbb{R}$ to $\mathbb{R}$. The achieved scalar values $H(\mathbf{X_i(t)}, \ve{\beta})$ are supplied into subsequent layers of \textit{numerical neurons} (e.g., the inputs and outputs are both scalar values) for further manipulations till the output layer that holds the response variable. An example FNN with three functional neurons on the first layer and two numerical neurons on the second layer is given in Fig \ref{FNN_exp}. 

FNN can be trained with the gradient descent approach under certain assumptions \cite{rossi2002functional, wang2019multilayer}. The \textit{forward propagation} step can go through as follows. First, in the functional neurons, the integral $\int_{t\in \mathcal{T}}W_{r}(\ve{\beta}_{r},t)X^{(i,r)}(t)dt$ is approximated by the numerical integration techniques. The output value $H(\mathbf{X_i(t)}, \ve{\beta})$ can then calculated by the formula in Eq (\ref{e2.2.1}). The forward propagation calculation in subsequent numerical layers is straightforward. In the \textit{backward propagation} step, the partial derivatives from the output layer up to the second hidden layer (i.e., the numerical layer after the functional neuron layer) can be easily calculated as in the classic neural networks. Whereas, it is essential to ensure that the partial derivatives of the values at the second layer (i.e., $H(\mathbf{X_i(t)}, \ve{\beta}))$ with respect to the parameters $\ve{\beta}_{r}$ exist. This requires that $\partial W_{r}(\ve{\beta}_{r},t)/\partial \beta_{r,q}$ exists almost everywhere for $t\in \mathcal{T}$. Under this assumption, $\partial H(\mathbf{X_i(t)}, \ve{\beta})/\partial \beta_{r,q}$ for any $r=1,..,R; q=1,...,Q_r$ can be estimated using numerical approximations of the following quantity
\begin{multline*}
\frac{\partial H(\mathbf{X_i(t)}, \ve{\beta})}{\partial \beta_{r,q}}=U^\prime(b + \sum_{r^\prime=1}^{R} \int_{t\in \mathcal{T}}W_{r}(\ve{\beta}_{r},t)X^{(i,r^\prime)}(t)dt)\\
\times \int_{t\in \mathcal{T}}\frac{\partial W_{r}(\ve{\beta}_{r},t)}{\partial \beta_{r,q}}X^{(i,r)}(t)dt,
\end{multline*} 
where $U^\prime(\cdot)$ represents the first derivative of the activation function $U(\cdot)$.

Note that the FNN described above requires dense regular covariates, i.e., there are a large number of data in $\mathbf{X^{(i,r)}}$ and the gaps among observation times $\mathbf{T^{(i,r)}}$ are relatively small, for $r=1,...,R; i=1,...,N$ \cite{rossi2002functional}. To handle scenarios where the accessible observations are sparse, an FNN equipped with a sparse functional neuron is proposed in \cite{wang2019multilayer, wang2017two}. In this paper, we focus on the FNN for dense data in Fig \ref{FNN_exp} and Eq (\ref{e2.2.1}). Nevertheless, we expect that our proposals work for the sparse FNN in \cite{wang2019multilayer} analogously. 

Although FNN has proven powerful for solving the temporal dependencies in temporal regression problems, it cannot appropriately handle spatial heterogeneity and dependencies in spatio-temporal data. In the following two sections, we propose two extensions of FNN to solve the spatio-temporal regression problem defined in Section \ref{sec2.1}.

\section{Geographically Weighted Functional Neural Network.}\label{sec3}
In this section, we present our Geographically Weighted FNN (GWFNN). The key idea is to propose an extension of FNN that is equipped with the geographically weighted regression technique to learn location-specific regression functions in Eq (\ref{e1.22}) \cite{yamanishi20038}. In particular, GWFNN is designed to not only handle the spatially varying relationships in the spatio-temporal regression problem, but also to implicitly consider the interactions among subjects through the geographical law-based weighting strategy. In this section, we describe the kernel weight functions used to weigh the raw data samples, the training of GWFNN, and the application of the learned model to new data. 


\subsection{Kernel Weight Functions.}\label{sec3.1}
\begin{figure*}
	\centering
	\begin{subfigure}[t]{1.78in}
		\centering
		\includegraphics[width=46mm]{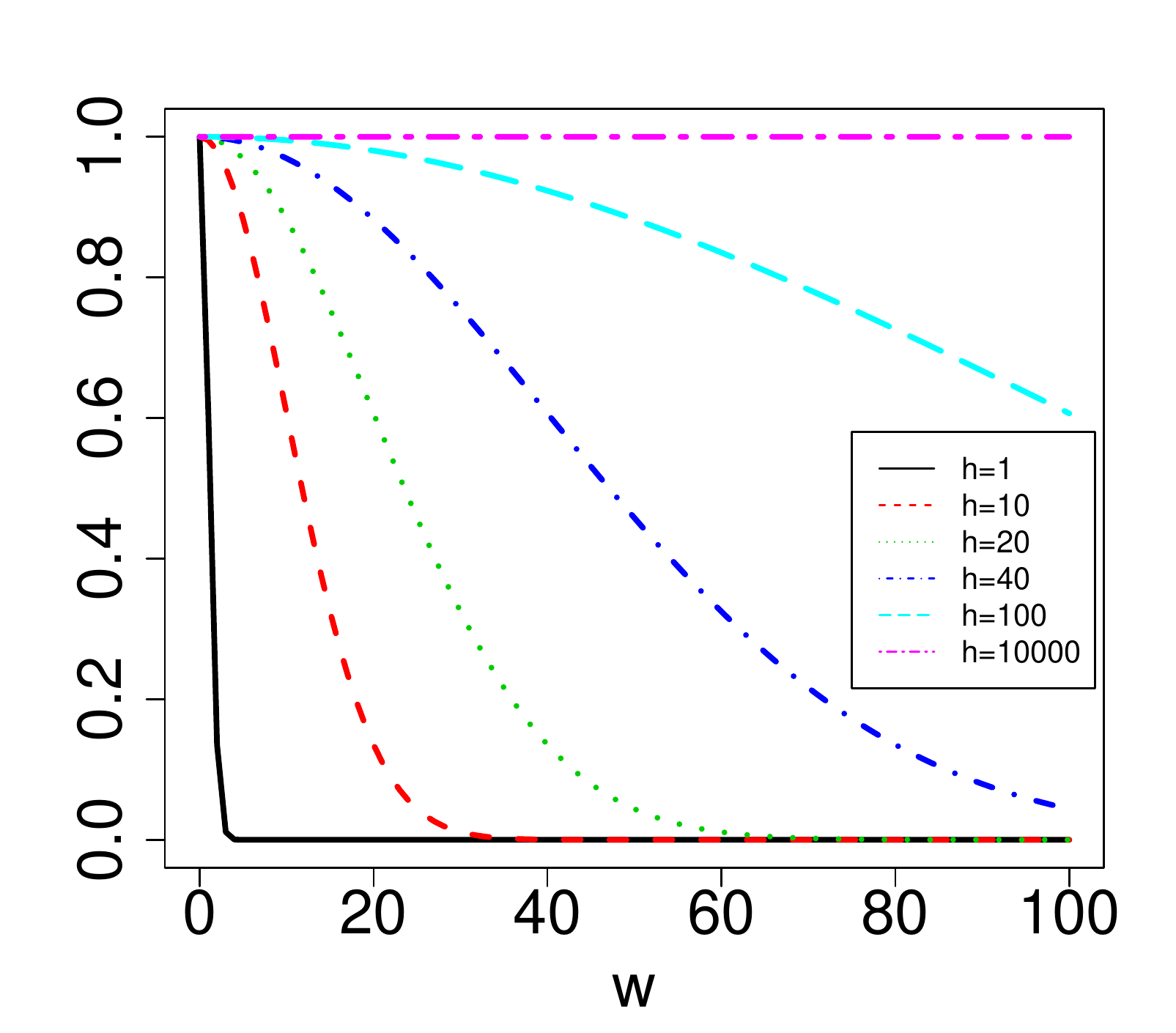} \label{k1}	
		\vspace{-0.15in}
		\caption{Gaussian}
	\end{subfigure}
	\quad
	\begin{subfigure}[t]{1.78in}
		\centering
		\includegraphics[width=46mm]{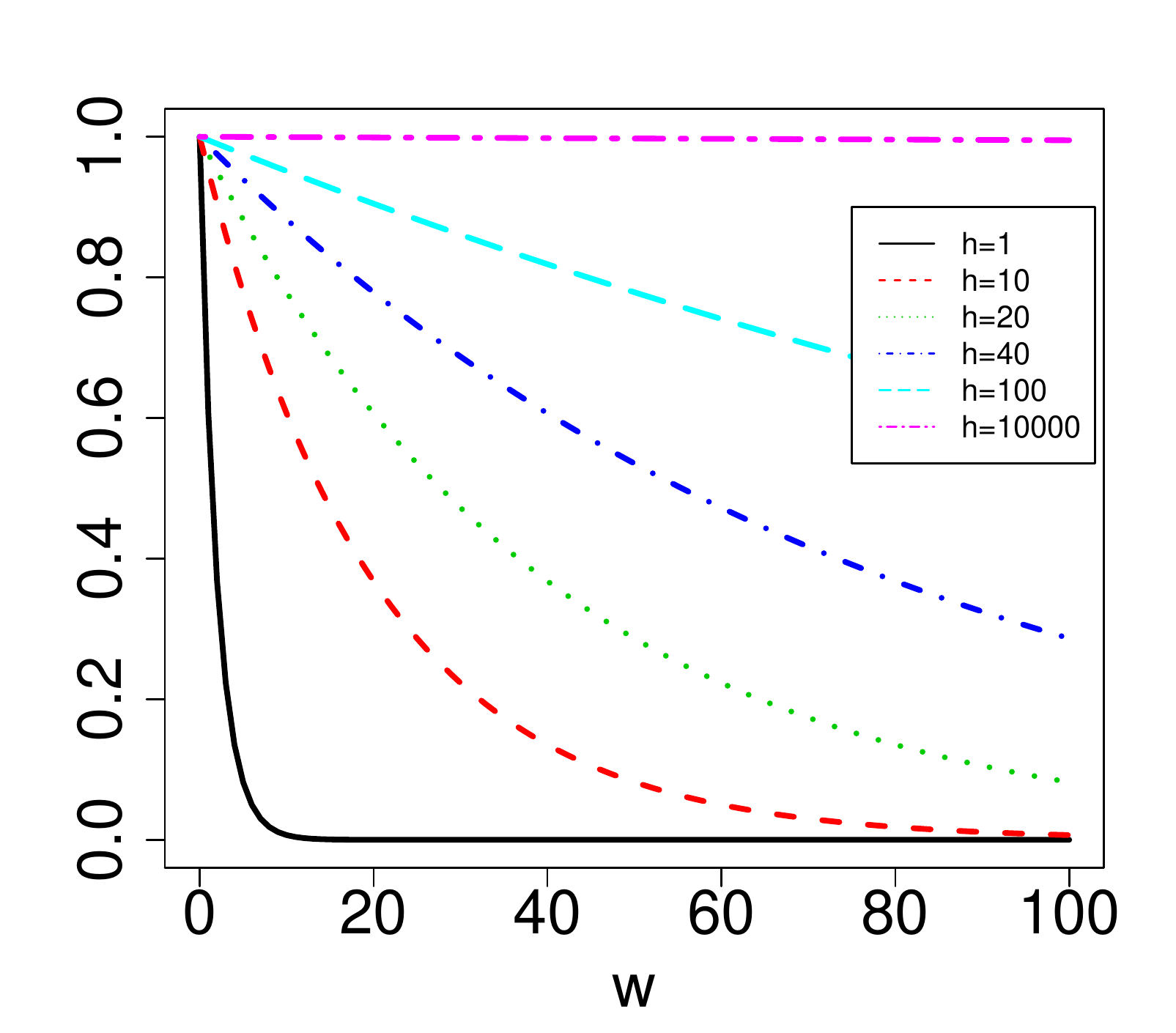} 	\label{k2}	
		\vspace{-0.15in}
		\caption{Exponential}
	\end{subfigure}
	\quad
    \begin{subfigure}[t]{1.78in}
		\centering
		\includegraphics[width=46mm]{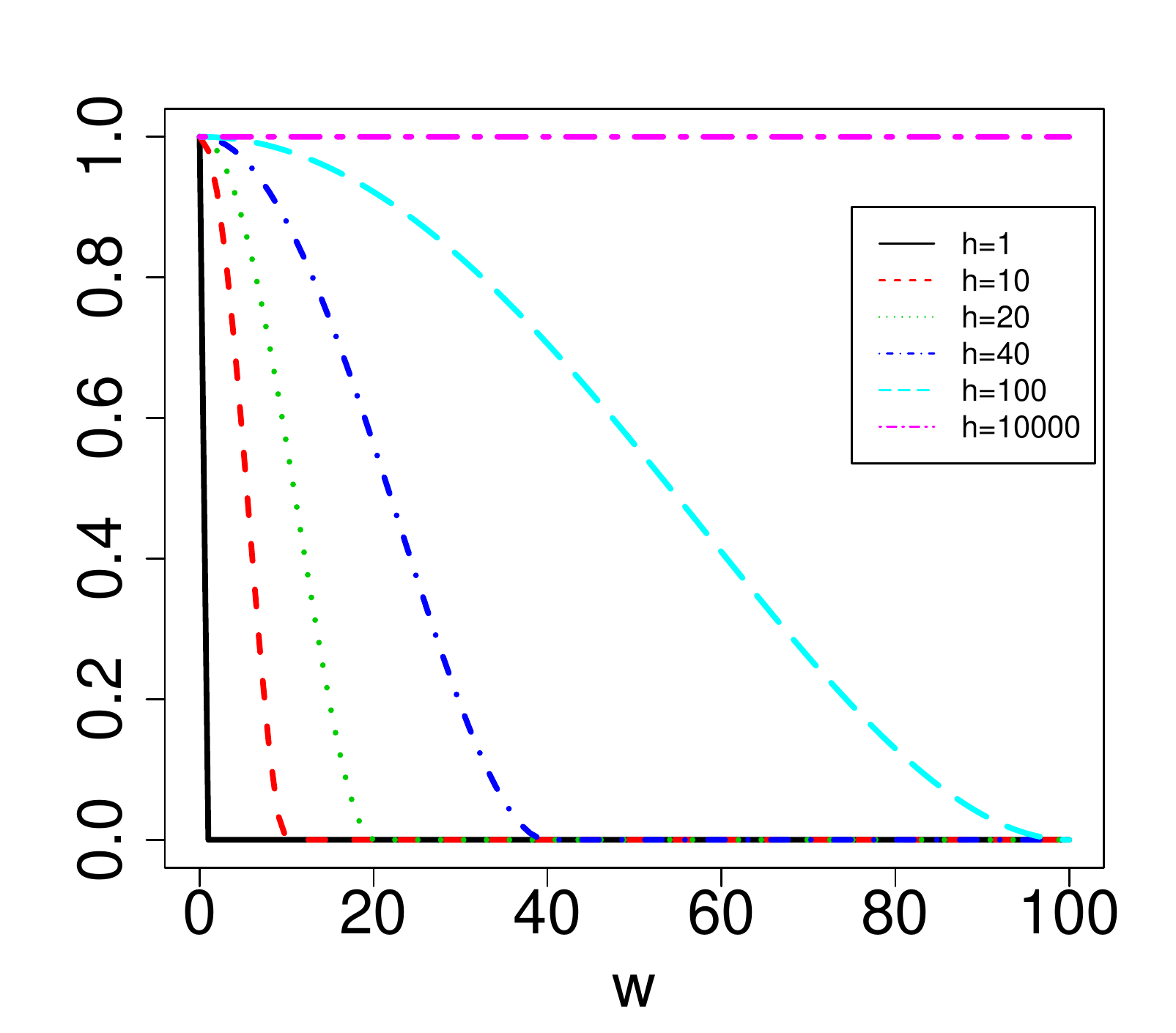} 	\label{k3}	
		\vspace{-0.15in}
		\caption{Double power}
	\end{subfigure}
	\caption{Kernel weight functions with different bandwidth $h$.}\label{kernel}
\end{figure*}

The functionality of the kernel weight function is to place higher weights on data samples that are potentially more similar to the considered regression point $u\in\{1,...,N\}$. According to the first law in geography which says that ``everything is related to everything else, but near things are more related than distant things'', we propose to use the Euclidean distance to quantify the similarity between samples. The distance between the $i$-th subject and the target $u$ is  
\begin{equation} \label{e3.1.1}
w_{i,u} = \norm{\mathbf{S_i}-\mathbf{S_u}}=\sqrt{(S_{i,1}-S_{u,1})^2+...+(S_{i,d}-S_{u,d})^2}.
\end{equation}
Theoretically, the kernel weight function can be any non-increasing functions of the Euclidean distance $w$. In this paper, we implement three widely-used kernel functions shown in Eq (\ref{e3.1.2}) and Fig \ref{kernel}. 

\begin{equation} \label{e3.1.2}
\begin{split}
K_{\text{Gaussian}}(w)& = e^{-0.5 (\frac{w}{h})^2}\\
K_{\text{Expo}}(w) &= e^{-(\frac{0.5w}{h})}\\
   K_{\text{DoublePower}}(w) &= \left \{
  \begin{aligned}
    &({1-(\frac{w}{h})^2})^2, && w < h \\
    &0, && w \geq h
  \end{aligned} \right.
\end{split}
\end{equation}


In the kernel functions, $h > 0$ is a bandwidth hyper-parameter that determines the rate at which the weights decay around a regression point $u$. Larger bandwidths lead to slower weight decays. When the bandwidth goes to infinity, the model will converge into a global regression where the relationship is homogeneous over space. In this sense, FNN is actually a special case of GWFNN with the same weights for all data points.

Choosing the most appropriate kernel function is a challenging problem. The optimal kernel function varies with the specific data properties. In practice, it is advisable to try multiple kernel functions and choose the best one based on the performance. For a given kernel function, the bandwidth parameter $h$ has been observed to influence the model performance. We propose to use the grid search and cross-validation to tune this hyperparameter.

\subsection{Training of GWFNN.}\label{sec3.2}
To facilitate the description of GWFNN, let's first briefly review the geographically weighted linear regression \cite{brunsdon1998geographically}. In i.i.d. linear regression models with $R$ scalar covariates, let $\mathbf{Z}$ be a $N\times (R+1)$ matrix containing the observed covariates, $\ve{\beta}$ be a $(R+1)$ dimensional vector containing the unknown parameters, and $\mathbf{Y}$ be a $N$ dimensional vector holding the responses. Then the unknown parameters are estimated by the least square estimator $\ve{\hat{\beta}}$ in Eq (\ref{e3.1.5}). 
\begin{equation} \label{e3.1.5}
\ve{\hat{\beta}} = (\mathbf{Z}^T\mathbf{Z})^{-1}\mathbf{Z}^T\mathbf{Y}
\end{equation}
When the parameters vary across locations, for a given target $u$, geographically weighted models first calculate the impact of each data sample on learning $\ve{\beta}_{\mathbf{S_u}}$, denoted as $\{W_{i,u}\}_{i=1}^N$. The location-specific estimator $\ve{\hat{\beta}}_{\mathbf{S_u}}$ is given in Eq (\ref{e3.1.6}). 
\begin{equation} \label{e3.1.6}
\begin{split}
\ve{\hat{\beta}}_{\mathbf{S_u}} & = (\mathbf{Z}^T \mathbf{W_u}\mathbf{Z})^{-1}\mathbf{Z}^T \mathbf{W_u}\mathbf{Y}\\
\mathbf{W_u}&=\text{diag}(W_{1,u},...,W_{N,u})
\end{split}
\end{equation}
Comparing Eq (\ref{e3.1.5}) and Eq (\ref{e3.1.6}), it can be seen that geographically weighted linear regression models are equivalent to the linear regression models with covariate matrix $\mathbf{W_u}^{1/2}\mathbf{Z}$ and response vector $\mathbf{W_u}^{1/2}\mathbf{Y}$, where  $\mathbf{W_u}^{1/2} \ve{\times} \mathbf{W_u}^{1/2} = \mathbf{W_u}$, with `$\ve{\times}$' being the matrix multiplication. 

This observation motivates us to train GWFNN as follows. For any target subject $u$, we specify the kernel weight function and a candidate set of $h$. For a given $h$, we use the corresponding kernel function formula to calculate the weights $W_{1,u},...,W_{N,u}$. Next, we propose to manipulate the raw temporal covariates $\mathbf{X_i(t)}=[X^{(i,1)}(t),...,X^{(i,R)}(t)]^T$ and the response $Y_i$ for $i=1,...,N$ by
\begin{equation} \label{e3.1.7}
\begin{split}
\mathbf{\tilde{X}_i(t)} &= \sqrt{W_{i,u}}\mathbf{X_i(t)}\\
\tilde{Y}_i&= \sqrt{W_{i,u}}Y_i.\\
\end{split}
\end{equation}
Then the transformed data $\mathbf{\tilde{X}_i(t)}$ and $\tilde{Y}_i$ that reflects the variety in the data importance is supplied into the FNN to achieve subject $u$'s estimator $\ve{\hat{\theta}}_{\mathbf{S_u}, h}$ for the considered $h$. The selection of $h$ will be discussed after we present the application phase of the learned model in the next subsection.

\subsection{Applying GWFNN to New Data.}\label{sec3.3}
Let the covariates be $\mathbf{X_{\text{new}}(t)}$ and the spatial location be $\mathbf{S_{\text{new}}}$ for a new data sample. Given the learned model, the objective is to predict $Y_{\text{new}}$. 

There are two possible scenarios when applying the trained model. The first is
the \textit{in-sample prediction} \cite{goulard2017predictions}, i.e., $\mathbf{S_{\text{new}}} \in \{\mathbf{S_1},..., \mathbf{S_N}\}$. For instance, a trained precipitation prediction model based on the weather data of 34 Chinese cities in 2005 is used to predict their annual precipitations in 2006 using the daily temperature data \cite{huang2018spatial}. The second is called the \textit{out-of-sample prediction} \cite{goulard2017predictions}, i.e., $\mathbf{S_{\text{new}}} \notin \{\mathbf{S_1},..., \mathbf{S_N}\}$. Out-of-sample predictions are necessary for a lot of spatial data analysis, where nascent temporal covariates data become gradually accessible. A real case study that demonstrates this phenomenon is discussed in \cite{goulard2017predictions}.

Predicting for in-sample data is straightforward. In the training phase in Section \ref{sec3.2}, the parameters for the $N$ existing locations have been learned, denoted as $\ve{\hat{\theta}}_{\mathbf{S_u}, h}$, for $u=1,...,N$. When new covariates $\mathbf{X_{\text{new}}(t)}$ are observed at one of the $N$ locations, the response variable can be predicted by plugging the new covariates into the learned model at the corresponding spatial location.

Unlike in-sample predictions, the parameter $\ve{\theta}_{\mathbf{S_{\text{new}}}, h}$ in out-of-sample scenarios is unknown and needs to be estimated. Under the assumption that the mapping $F(\cdot)$ at the new location can be approximated by interpolating the mappings at the training sites, we proceed as follows to obtain the prediction. 
\begin{enumerate}[leftmargin=*]
    \itemsep0em 
    \item Use the selected kernel function to calculate the weights of the $N$ training data points with respect to the new data point based on their distance to $\mathbf{S_{\text{new}}}$.  The achieved weights are denoted as $W_{1,\text{new}},...,W_{N,\text{new}}$.
    \item Obtain transformed training data $\{\sqrt{W_{i,\text{new}}}\mathbf{X_i(t)}, \sqrt{W_{i,\text{new}}}Y_i\}_{i=1}^{N}$.
    \item Train FNN with transformed data to obtain the unknown parameter for this new location, denoted as $\ve{\hat{\theta}}_{\mathbf{S_\text{new}}, h}$.
    \item Given $\ve{\hat{\theta}}_{\mathbf{S_\text{new}}, h}$, predict the response by plugging in the covariates $\mathbf{X_{\text{new}}(t)}$. 
\end{enumerate}
Note that the parameter learning procedure is different from the training step where the data at the target location and the remaining locations are both used to learn the target parameters. In this phase, it is infeasible to include the target location (i.e., the new location) because the response for this point is unknown. 

Based on the above discussion on predictions, the selection of the bandwidth parameter $h$ can be straightforwardly conducted through cross-validations.

\section{Spatial Autoregressive Functional Neural Network.}\label{sec4}
In this section, we present the proposed Spatial Autoregressive FNN (SARFNN). We propose an extension of FNN that includes spatial autoregressive terms in the model (see Eq (\ref{e1.33})). SARFNN is specially constructed to account for the impact of nearby location's response values. It is noteworthy that it implicitly handles the spatial heterogeneity, as shown by the experiments.


\subsection{Training of SARFNN. }\label{sec4.1}
Following the formulation in the existing spatial autoregressive (SAR) predictive models \cite{huang2018spatial, cressie2015statistics}, the main intuition is that the response at a target location within $\mathcal{D}$ is partially determined by the responses at nearby locations, in addition to the influences of the local temporal predictors. The strength of the spatial correlation follows the first law of geography, i.e., it decreases with the increasing of the Euclidean distances. Quantitatively, the kernel function should be a non-increasing function of the Euclidean distance $w$ when $w>0$ and takes value 0 at $w=0$. This is because we should avoid using the response of the target location itself. The new kernel functions achieved by setting the value of functions in Eq (\ref{e3.1.2}) be 0 at $w=0$ are respectively denoted as $\tilde{K}_{\text{Gaussian}}(w)$, $\tilde{K}_{\text{Expo}}(w)$, $\tilde{K}_{\text{DoublePower}}(w)$. Another extensively-used kernel function in the literature is  
\begin{equation} \label{e4.1.1}
\tilde{K}_{\text{Nearest}}(w) = \left \{
  \begin{aligned}
    &1, && 0 < w \leq w_{i,u (h)} \\
    &0, && \text{otherwise},
  \end{aligned} \right.
\end{equation}
where $w_{i,u (h)}$ is the $h$-th order statistics of the Euclidean distances $w_{1,u},...,w_{N,u}$. The kernel function in Eq (\ref{e4.1.1}) sets the weight of locations with the $h$ smallest distances to $u$ as $1$ and $0$ otherwise. Note that $h$ is a bandwidth parameter to be chosen through cross-validations, similar to the strategy in Section \ref{sec3}. When the kernel functions take value 0 for $w\geq 0$, SARFNN converges to FNN. In this sense, FNN is a special case of SARFNN with zero weights for all data points.

Let $\mathbf{Y}=[Y_{1},...,Y_{N}]^T$, and $\mathbf{\tilde{W}}^*$ be a $N\times N$ weight matrix whose $(i,i^\prime)$'s element is $\tilde{K}(w_{i,i^\prime})$. $\mathbf{\tilde{W}}^*$ is standardized to obtain a row-normalized matrix $\mathbf{\tilde{W}}$, whose row summations are 1. Let $\mathbf{X_i(t)}=[X^{(i,1)}(t), ..., X^{(i,R)}(t)]^T$ and $\mathbf{\tilde{W}_i}$ be the $i$-th row of $\mathbf{\tilde{W}}$. Based on the data samples, we proposed to train SARFNN by learning the homogeneous mapping from the temporal covariates $\mathbf{X_i(t)}$ and the spatial autoregressive covariate $\mathbf{\tilde{W}^T_i}\mathbf{Y}$ to the response $Y_i$, for $i=1,...,N$.

\begin{figure}
\centering
\includegraphics[width=85mm]{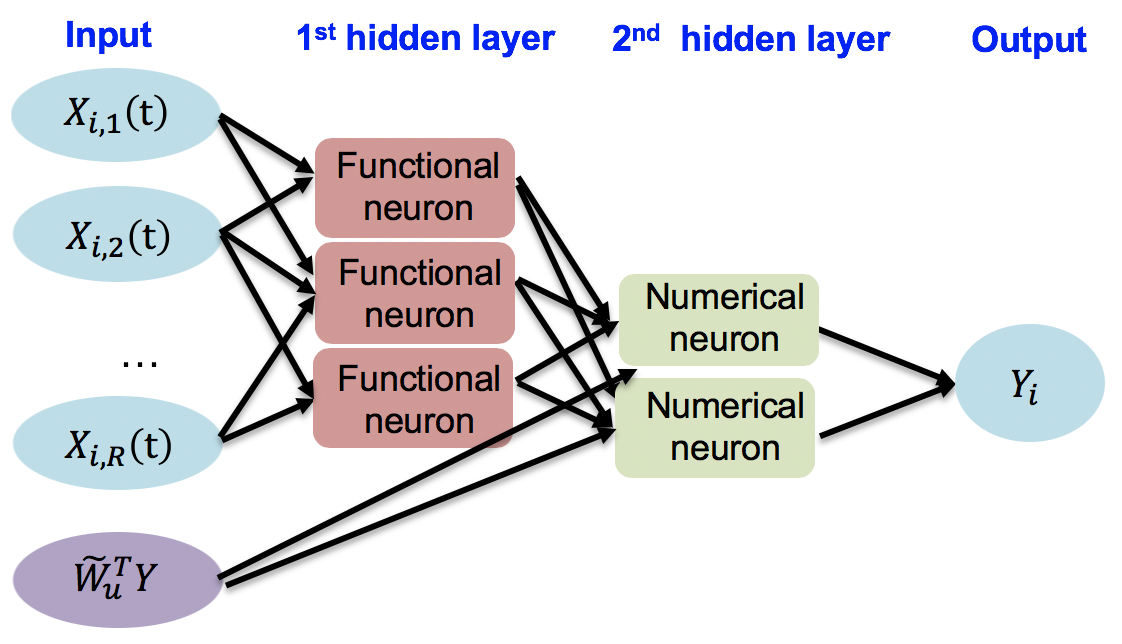}
  \caption{An example of the proposed SARFNN, where there are three functional neurons on the first layer and two numerical neurons on the second layer.}
  \label{SARFNN_exp}
\end{figure}

To build a network with both the temporal covariates and the scalar spatial covariate, we proposed to separately consider the two types of inputs on the first layer. The temporal covariates are transformed in the same manner as the FNN in Fig \ref{FNN_exp}, while no operation is done on the scalar spatial autoregressive covariate. Next, the obtained data are fed into subsequent fully connected numerical layers. An example of our proposed SARFNN is visualized in Fig \ref{SARFNN_exp}. The proposed SARFNN can be analogously trained by the gradient descent approach described in Section \ref{sec2.2}. 

\subsection{Applying SARFNN to New Data.}\label{sec4.2}
Let the new $R$-dimensional temporal covariates be denoted by  $\mathbf{X_{\text{new}}(t)}$ and the new location be $\mathbf{S_{\text{new}}}$. In the application phase, the purpose is to predict the corresponding response $Y_{\text{new}}$. For the proposed SARFNN model, no additional parameter estimation is needed, as the same set of parameter is used across all spatial locations. 

First, we consider the in-sample predictions, i.e., $\mathbf{S_{\text{new}}} \in \{\mathbf{S_i}\}_{i=1}^N$. Suppose that $\mathbf{S_{\text{new}}}$ the same as the $l$-th subject in the training, the response variable can be estimated by feeding $\mathbf{X_{\text{new}}(t)}$ and $\mathbf{\tilde{W}^T_{l}} \mathbf{Y}$ into the learned SARFNN.


Next, we describe our proposal for the out-of-sample predictions, i.e., $\mathbf{S_{\text{new}}} \notin \{\mathbf{S_i}\}_{i=1}^N$. Following the approach in \cite{goulard2017predictions}, we propose to calculate the kernel values of the training samples with respect to the new location, i.e., $\tilde{K}(w_{i,new})$, for $i=1,...,N$, and store them in a $N$-dimensional vector $\mathbf{\tilde{W}}^*_{oS}$. Then a $(N+1)\times (N+1)$ weight matrix that includes both the training data and the new data is given in Eq (\ref{e4.1.4}). 
\begin{equation} \label{e4.1.4}
\mathbf{\tilde{W}}^*_{\text{new}} = \begin{bmatrix}
    \mathbf{\tilde{W}}^*       & \mathbf{\tilde{W}}^*_{oS}\\
    (\mathbf{\tilde{W}}^{*}_{oS})^T      & 0 
\end{bmatrix}
\end{equation}
Note that $\tilde{W}^*$ is the $N\times N$ matrix for the training data. The row-normalized matrix of $\mathbf{\tilde{W}}^*_{\text{new}}$ is denoted as $\mathbf{\tilde{W}}_{\text{new}}$. After matrix transformations, suppose that the $N$-dimensional vector $\mathbf{\tilde{W}}^*_{oS}$ in $\mathbf{\tilde{W}}_{\text{new}}$ becomes vector $\mathbf{\tilde{W}}_{oS}$, our out-of-sample prediction is achieved by plugging the temporal covariates $\mathbf{X_{\text{new}}(t)}$ and $\mathbf{\tilde{W}}^T_{oS}\mathbf{Y}$ to the trained SARFNN.

\section{Numerical Experiments.}\label{sec5}
In this section, three numerical experiments are conducted to demonstrate the effectiveness of our proposed models, in comparison with other feasible baselines listed in Table \ref{base}. Two simulation studies are first considered to show the performances of the proposed models under two different scenarios, i.e., data that exhibits either heterogeneity or dependencies. The considered models are then applied to solve a real-world challenge in the meteorology field.


\begin{table}
\center
\caption {Baselines considered in the experiments.}  \label{base} 
\begin{tabular}{l|l}
\hline
\hline
\textbf{Baseline}    & \textbf{Description}  \\ \hline
FLM & The conventional i.i.d functional linear regression model \\
 & in Eq (\ref{flm}).\\
$\text{FLM}_{\text{SP}}$ & Treat the spatial information $\mathbf{S_i}$ as additional scalar \\
 & covariates in FLM. \\
GWFLM & Geographically weighted functional linear model \cite{yamanishi20038}.\\
SARFLM & The spatial autogressive functional linear model \cite{huang2018spatial}.\\
FNN & The traditional i.i.d functional neural network \cite{rossi2002functional}.\\
$\text{FNN}_{\text{SP}}$ & Treat the spatial information $\mathbf{S_i}$ as additional scalar\\
 & covariates in FNN. \\
\hline
\hline
\end{tabular}
\end{table}


\subsection{Simulation I: Data with Spatial Heterogeneity.}\label{sec5.1} 
\begin{figure}
	\centering
	\begin{subfigure}[t]{1.6in}
		\centering
		\includegraphics[width=42mm]{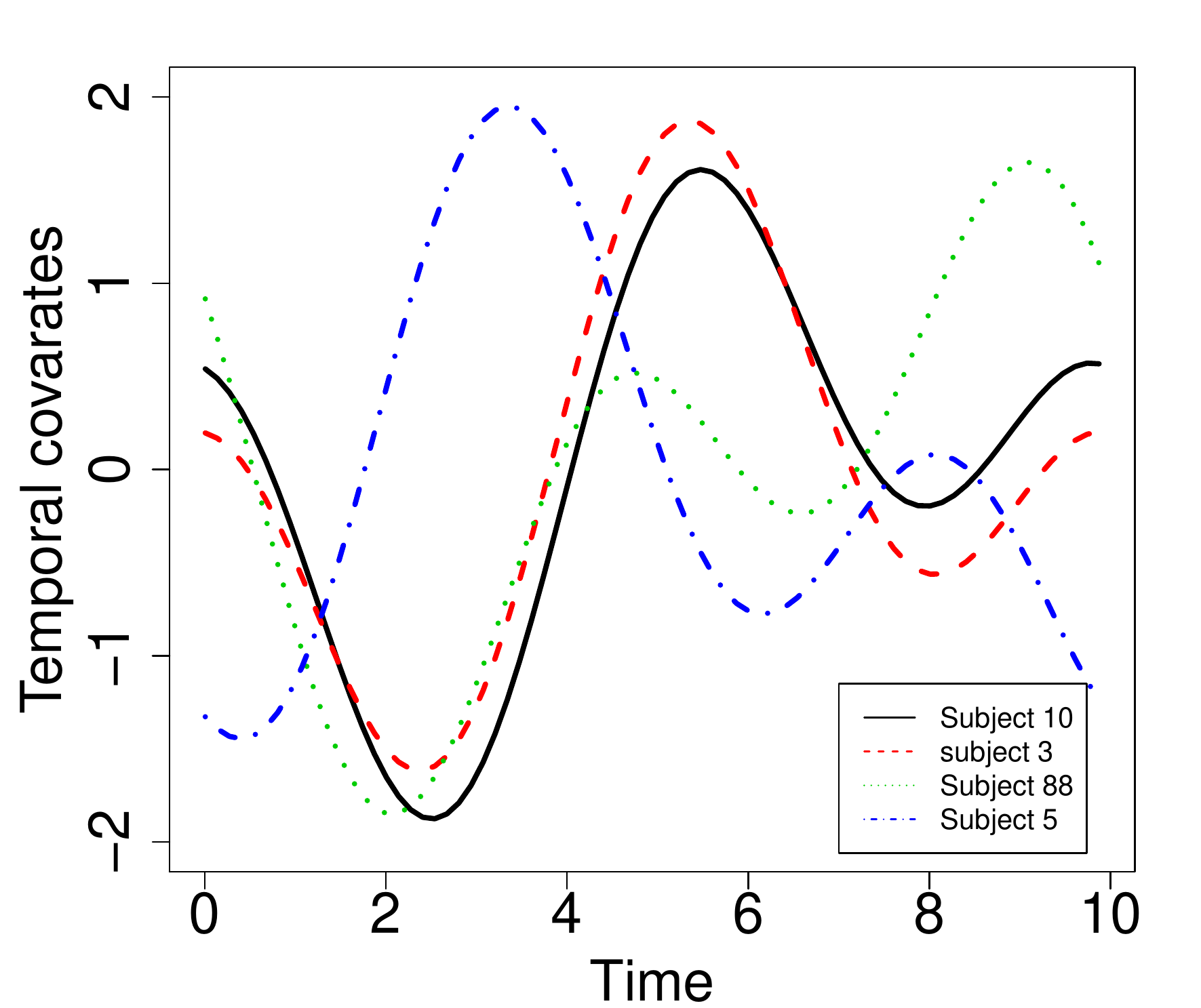} 
		\vspace{-0.15in}
		\caption{Examples of $X^{(i)}(t)$.}\label{x_exp}	
	\end{subfigure}
	\quad
	\begin{subfigure}[t]{1.6in}
		\centering
		\includegraphics[width=42mm]{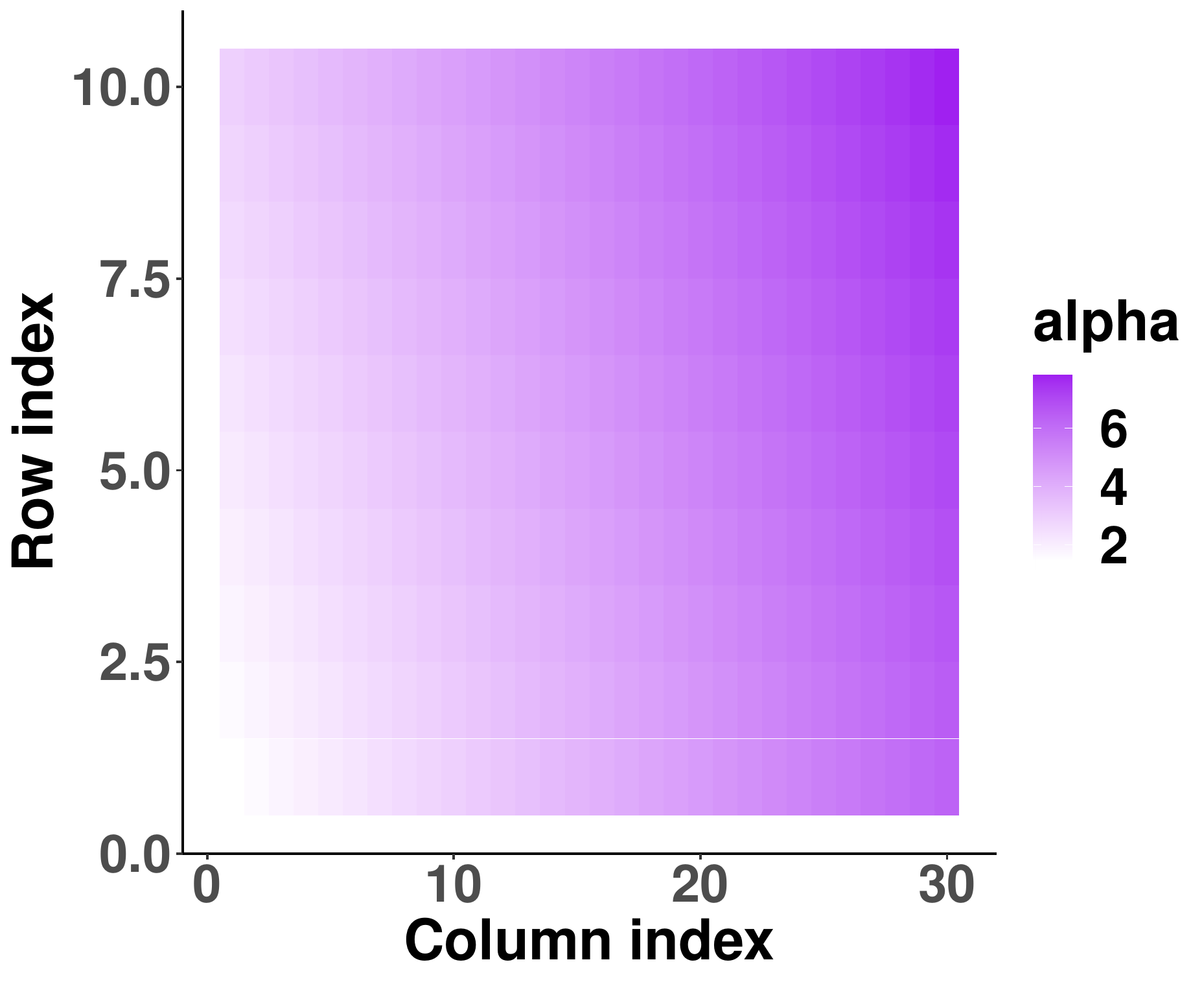} 	
		\vspace{-0.15in}
		\caption{Value of $\alpha_i$.}\label{gwr_theta}	
	\end{subfigure}
	\caption{Visualizations for Simulation I. }\label{sim1_plot}
\end{figure}

In this simulation study, we consider a scenario where the ground truth relationship between the temporal covariates and the response is a non-linear mapping with spatially varying parameters. The setting is motivated by the simulation studies in \cite{yang2014extension, muller2013continuously}. The spatial space is assumed to be a grid of regular square cells with $P$ rows and $Q$ columns, as shown in Fig \ref{gwr_theta}. Each subject is located in one of the $P\times Q$ cells. We set $P=10$ and $Q=30$, then the sample size is $N=10\times30=300$. The $i$-th subject is located at the $(\floor{i/30}+1)$-th row and the column index is the remainder of $i/30$. The row and column indexes for subject $i$ are denoted as $p_i$ and $q_i$ respectively. 

We consider a case where $R=1$, i.e., there is one temporal covariate. The considered time range is $\mathcal{T}=[0, 10]$. The temporal covariates are generated by $X^{(i)}(t)=\sum_{k=1}^{4}\xi_{i,k}\phi_k(t)$, with $\phi_1(t)=\sqrt{2}\sin(2\pi t)$, $\phi_2(t)=\sqrt{2}\cos(2\pi t)$, $\phi_3(t)=\sqrt{2}\sin(4\pi t)$, and $\phi_4(t)=\sqrt{2}\cos(4\pi t)$. The coefficients are $\xi_{i,1}=\cos(U_i)$, $\xi_{i,2}=\sin(U_i)$, $\xi_{i,3}=\cos(U^\prime_i)$, $\xi_{i,4}=\sin(U^\prime_i)$, with $U_i, U^\prime_i$ being independent samples of Uniform$[0,2\pi]$, for $i=1,...,N$. Visualizations of the temporal covariate for four randomly selected subjects are given in Fig \ref{x_exp}. The response variable is generated by formula
\begin{equation} \label{e5.1.1}
Y_{i} = \alpha_i + \int_{0}^{10}\cos(t-X^{(i)}(t)-5)\,dt + \epsilon_i, 
\end{equation}
where $\epsilon_i \sim$ i.i.d $N(0, 0.5^2)$. Note that $\alpha_i$ is the spatially variant parameter whose value is similar among nearby locations in the $10\times 30$ cell grid. Specifically, it is a function of the row and column indexes ($p_i$, $q_i$) with formula
\begin{equation} \label{e5.1.2}
\alpha_i = 1 + \frac{p_i+q_i}{6}.
\end{equation}
The value of $\alpha_i$ over the cell grid is shown in Fig \ref{gwr_theta}.

The bandwidth parameter $h$ for spatial functional models are selected by cross-validation. A popularly-used parameter function $W(\ve{\beta},t)$ is assumed across all methods, 
\begin{equation} \label{e5.1.3}
W(\ve{\beta},t) = \sum_{k=1}^{K_{sel}} \beta_{k}\hat{\psi}_k(t), 
\end{equation}
where $\hat{\psi}_k(t)$ is the estimated eigenfunction of the covariance function of samples $\{X^{(i)}(t)\}_{i=1}^N$ from the functional principal component analysis \cite{ramsay2006functional,Reimherr2017FDA}, a functional counterpart of the PCA for vectors. The parameter function in Eq (\ref{e5.1.3}) is widely used in the FDA literature for a vast range of problems \cite{ramsay2006functional,Reimherr2017FDA, rossi2002functional, james2009functional}. Following the convention in FDA, $K_{sel}$ is selected using the fraction of variance explained approach with the commonly-used cutoff value $0.99$. In our experiments, the estimated $K_{sel}$ is mostly within the range of 3 to 6. 

The functional linear models are implemented using the R package `refund' \cite{refundR}. In FNNs, we specify a network structure with two hidden layers, including the first layer with four functional neurons and followed by a layer with two numerical neurons. The corresponding weight functions in FLMs and the first layer of FNNs have the format shown in Eq (\ref{e5.1.3}).  

The average root mean square errors (RMSE) over 10-fold cross-validations are given in the second column of Table \ref{sim1_tab}. Note that the predictions for new locations are obtained through the out-of-sample prediction strategies in Section \ref{sec3.3} and \ref{sec4.2}. The simulated relationship is non-linear, therefore all the FNN type methods are expected to perform better than the linear models. When calculating the improvement (`IMP' in Table \ref{sim1_tab}), we chose two different baselines, i.e., FLM for all the linear models and FNN for all the non-linear models. Here are the key observations. First, the FNN models have better performances than the FLM models. Second, $\text{FNN}_{\text{SP}}$ has slightly better performance than the conventional FNN that ignores the spatial information. Besides, our proposed models (GWFNN, SARFNN) outperforms all the other methods. The improvement achieved by GWFNN models (around $18\%$) is higher than the SARFNN model (around $11\%$), which justifies our intuition that GWFNN is more suitable for data that explicitly exhibit the spatial heterogeneity rather than dependencies. 

\begin{table*}
\center
\caption {Experimental results: 10-fold cross-validation results for Simulation I and II, and leave-one-out cross-validation results for Canada Weather Data.}  \label{sim1_tab} 
\begin{tabular}{cl||cc||cc||cc}
\hline
\hline
&  & \multicolumn{2}{c||}{\textbf{(Simulation I)}}   & \multicolumn{2}{c||}{\textbf{(Simulation II)} }  & \multicolumn{2}{c}{\textbf{(Real data)} } \\ 
\textbf{Category} & \textbf{Method} & \textbf{RMSE} & \textbf{IMP} & \textbf{RMSE} & \textbf{IMP} & \textbf{RMSE} & \textbf{IMP}\\ \hline
Linear & FLM   & 1.143   & -  & 1.182   & -  & 0.725   & -\\ 
       & $\text{FLM}_{\text{SP}}$  & 1.126  & $1.49\%$ & 1.186  & $-0.34\%$ & 0.708  & $2.34\%$  \\
       & $\text{GWFLM}_{\text{Gaussian}}$ & 1.141 & $0.17\%$ & 1.182 & $0.00\%$  & 0.536 & $26.07\%$\\ 
       & $\text{GWFLM}_{\text{Expo}}$ &  1.139  & $0.35\%$ & 1.182  & $0.00\%$  & 0.550  & $24.14\%$ \\
       & $\text{GWFLM}_{\text{DoublePower}}$ &  1.138  & $0.44\%$ & 1.182  & $0.00\%$ & 0.535  & $26.21\%$\\ 
       & $\text{SARFLM}_{\text{Nearest}}$ & 1.144  & $-0.09\%$ & 1.179 & $0.25\%$ & 0.664  & $8.41\%$ \\ 
\hline
Non-linear & FNN  & 0.564 & -  & 0.705 & -  & 0.740 & - \\ 
           & $\text{FNN}_{\text{SP}}$   & 0.521  & $7.62\%$  & 0.709  & $-0.57\%$ & 0.671  & $9.32\%$ \\ 
           & $\text{GWFNN}_{\text{Gaussian}}$ & \textbf{0.461} & $\boldsymbol{18.26\%}$ &0.677 & $3.96\%$ & \textbf{0.467} & $\boldsymbol{36.89\%}$  \\ 
           & $\text{GWFNN}_{\text{Expo}}$ & 0.504 & $10.64\%$  & 0.711 & $-0.85\%$ &  0.594 & $19.73\%$ \\ 
           & $\text{GWFNN}_{\text{DoublePower}}$ &  \textbf{0.467} & $\boldsymbol{17.20\%}$ &  0.710 & $-0.71\%$ &  \textbf{0.494} & $\boldsymbol{33.24\%}$\\ 
           & $\text{SARFNN}_{\text{Nearest}}$  &0.502 & $11.00\%$  & \textbf{0.647} & $\boldsymbol{8.23\%}$  & 0.638 & $1.38\%$ \\ 
\hline \hline
\multicolumn{6}{l}{\footnotesize * IMP w.r.t FLM is ($\text{RMSE}_{\text{FLM}}-\text{RMSE})$/$\text{RMSE}_{\text{FLM}}$}\\
\multicolumn{6}{l}{\footnotesize * IMP w.r.t FNN is ($\text{RMSE}_{\text{FNN}}$-\text{RMSE})/$\text{RMSE}_{\text{FNN}}$}\\
\multicolumn{6}{l}{\footnotesize * `-' means not applicable}
\end{tabular}
\end{table*}


\subsection{Simulation II: Data with Spatial Dependencies.}\label{sec5.2}
In this study, we consider a setting where the response is dependent on those at nearby spatial locations. The structure of the two-dimensional space, the temporal covariates are generated in the same manner as Simulation I. The response variables $\mathbf{Y}=[Y_1,...,Y_N]^T$ are generated by 
\begin{equation} \label{e5.1.4}
\begin{split}
\tilde{Y}_{i} &= \int_{0}^{10}\cos(t-X^{(i)}(t)-5)\,dt + \epsilon_i\\
\mathbf{Y}&=(\mathbf{I}-0.25\mathbf{M})^{-1}\times \mathbf{\tilde{Y}}
\end{split}
\end{equation}
where $\tilde{Y}=[\tilde{Y}_{1},...,\tilde{Y}_{N}]^T$, $\epsilon_i \sim$ i.i.d $N(0, 0.5^2)$, $\mathbf{I}$ is the identify matrix, and $\mathbf{M}$ is the rook matrix \cite{huang2018spatial}, whose $(i,i^\prime)$ element $m_{i,i^\prime}=1$ if subject $i$ and $i^\prime$ are neighbours and $m_{i,i^\prime}=0$ otherwise. Note that cells in the grid that share at least an edge are defined as neighbours. 

The implementations of the models are the same as Simulation I. The average root mean square errors (RMSE) over 10-fold cross-validations are summarized in the third column of Table \ref{sim1_tab}. It can be seen that our proposed GWRFNN and SARFNN achieve the best results in terms of RMSE. The major observations are listed as follows. First, the FNN models have better performances than the FLM models. Second, all the spatio-temporal FLM models yield comparable accuracies with the conventional FLM models. Our explanation is that the assumed linear relationship in FLM models is inaccurate under the considered non-linear scenario such that adding spatio-temporal modeling techniques cannot improve the model performance. Third, the performance of GWFNN with different kernel functions is either slightly better (i.e., improvement of Gaussian kernel is $3.96\%$) or worse (the exponential and double power kernel functions), compared to the traditional FNN. This observation is amenable to our intuition that the effectiveness of GWFNN in handling the spatial dependency is highly dependent on the choice of kernel functions. Finally, SARFNN achieves the highest improvement $8.23\%$, as it explicitly considers the dependency among data. 

Given the observations in the two simulation studies, to achieve the best model performance, we recommend trying both GWFNN and SARFNN with multiple kernel functions configurations and deploying the model with the best performance. 





\subsection{Real Data: Annual Precipitation Prediction.}\label{sec5.3} 
Understanding the correlation between the evolution of daily average temperature over the year and the annual average precipitation is a topic of paramount importance in the meteorology field. Most of the meteorological measurements have associated geospatial information, such as longitude and latitude. If we view the daily average temperature over the year and the real-valued annual average precipitation as the corresponding input and output at each geospatial location, the considered problem naturally becomes a spatio-temporal regression problem defined in Section \ref{sec2.1}. 

\begin{figure}
	\centering
	\begin{subfigure}[t]{1.6in}
		\centering
		\includegraphics[width=42mm]{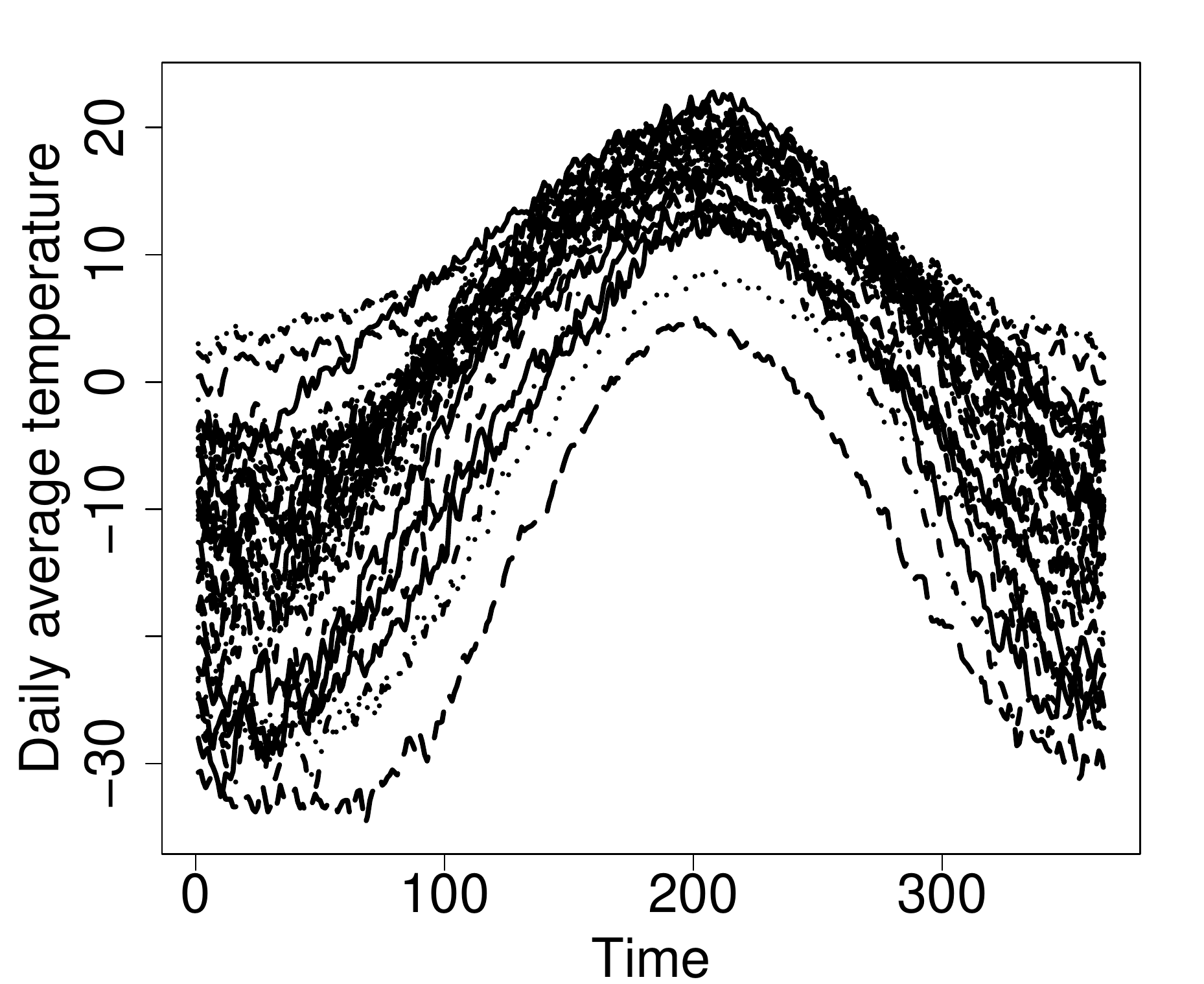} 
		\vspace{-0.15in}
		\caption{Visualizations of $X^{(i)}(t)$.} \label{cad1}	
	\end{subfigure}
	\quad
	\begin{subfigure}[t]{1.6in}
		\centering
		\includegraphics[width=42mm]{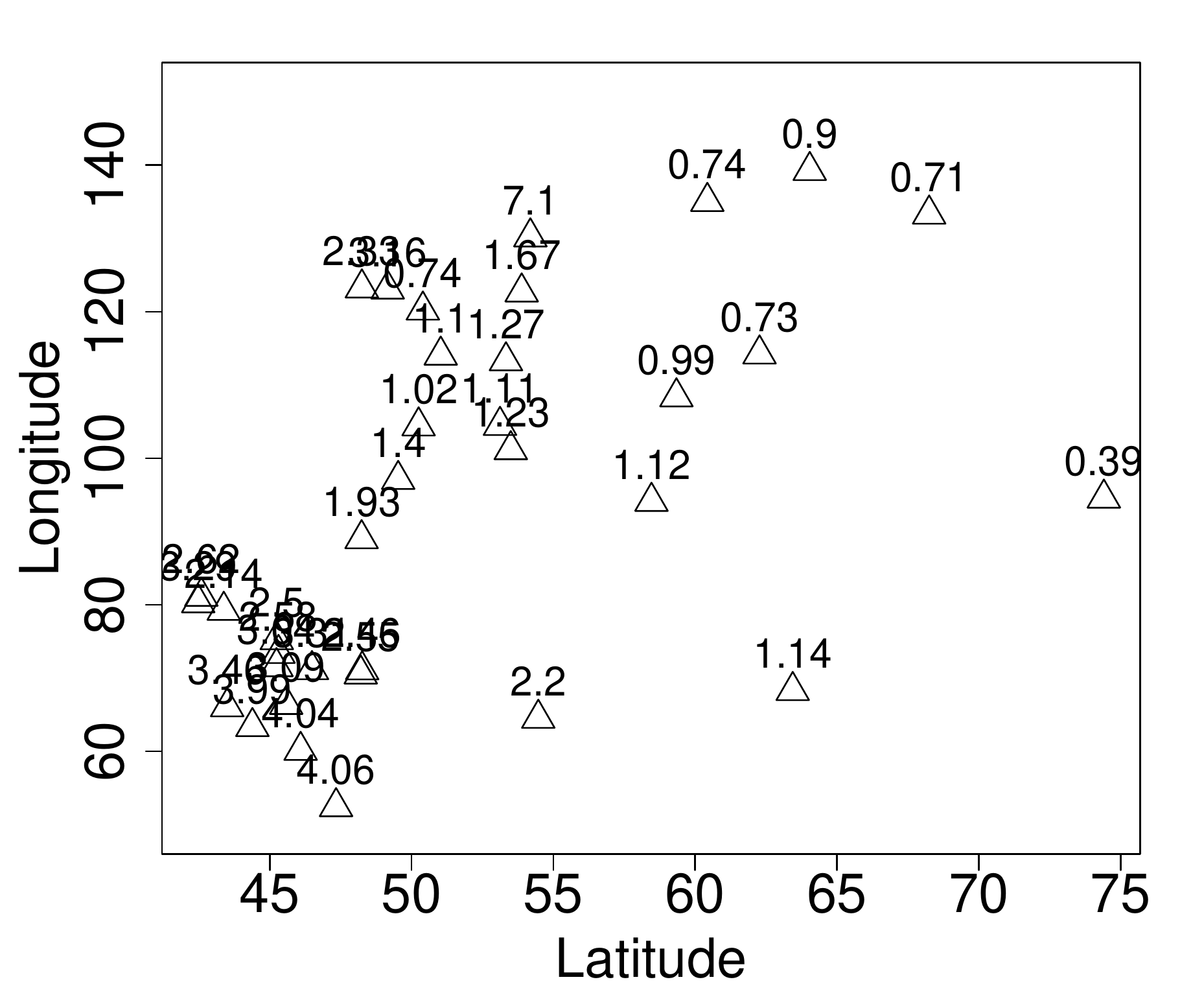} 	
		\vspace{-0.15in}
		\caption{Location and response.} 	\label{cad2}
	\end{subfigure}
	\caption{Visualizations for real-world data analysis. }\label{cad_plot}
\end{figure}

To demonstrate the effectiveness of our proposed models in solving the annual precipitation forecasting problem, we apply all the considered methods to the benchmark, the Canada weather data set. This data set is publicly available in the R package `fda' \cite{fdaR}. It contains the local meteorological measurements collected by 35 weather stations spreading out throughout Canada. The longitude and latitude of the weather stations are illustrated in Fig \ref{cad2}. For each weather station, the daily average temperature records over a specific year, i.e., a time series of length 365, are provided (see Fig \ref{cad1}). Also, for each station, the yearly average precipitation is given, as illustrated by the labels in Fig \ref{cad2}. Based on the $N=35$ samples, the objective is to build a regression model that leverages both temporal and spatial dependencies to minimize the prediction errors when using the daily temperature time series to predict the annual average precipitation. 

The implementations of the models are similar to Simulation I. The major difference is that the numbers of neurons in the two hidden layers of FNN are 8 and 4, respectively. As shown in the last column of Table \ref{sim1_tab}, our proposed GWFNN and SARFNN both achieve some improvement over the conventional FNN in terms of leave-one-out cross-validated RMSE. Especially, our proposed geographically weighted model achieves the smallest RMSE among all approaches, with a $30\%$ improvement over FNN. Note that we conducted statistical tests, including the T-test and Wilcoxon rank-sum test, to confirm that the improvements are statistically significant at significance level 0.05. The reason why the improvement of SARFNN is small is that the data probably only exhibit spatial heterogeneity instead of dependency. This is a reasonable articulation, as the 35 stations are so spread out that the data dependency is minimal.

\section{Conclusion and Discussion.}\label{sec6}
In this paper, we proposed two practical extensions of the i.i.d Functional Neural Network (FNN) to solve the critical spatio-temporal regression problem. The key idea behind the proposed models is to equip FNN with the geographically weighted regression and spatial autoregression techniques. Through meticulous simulation studies, we demonstrated the usefulness of the proposed models and investigated the advantages of each model for data with different spatial properties. Furthermore, we formulated the precipitation prediction in meteorology as a spatio-temporal regression problem and deployed the proposed models to tackle it. In all the numerical experiments, there exists a configuration under which the proposed models significantly outperformed the state-of-art methods. 

As discussed in the paper, choosing the most appropriate kernel function is a challenging problem and is out of the scope of the paper. The effectiveness of kernel functions varies with the underlying spatial properties in the data. In real practice, we recommend trying both GWFNN and SARFNN with multiple kernel function configurations and deploying the model with the best performance. We expect the proposed models provide additional useful insights to diverse real-world challenges that can be formulated as a spatio-temporal regression problem.


\balance

\bibliographystyle{IEEEtran}
\bibliography{fanova}

\end{document}